\documentclass[11pt]{article}

\usepackage[final]{acl}

\usepackage{times}
\usepackage{latexsym}
\usepackage[T1]{fontenc}
\usepackage[utf8]{inputenc}
\usepackage{microtype}
\usepackage{inconsolata}
\usepackage{mathtools}
\usepackage{amsmath, amssymb}
\usepackage{graphicx}
\graphicspath{{./}{./figures/}{./figures/no_intervention/}{./figures/intervention1/}{./figures/intervention2/}{./figures/intervention3/}{./figures/intervention4/}{./figures/intervention5/}{./figures/intervention5_2/}{./figures/obs1_appendix/}{./figures/obs2_appendix/}}
\usepackage{cleveref}
\usepackage{enumitem} 
\usepackage{pgfplots}
\pgfplotsset{compat=1.17}
\usepackage{caption}
\usepackage{subcaption} 
\usepackage{float}
\usepackage{xcolor}
\usepackage[bottom]{footmisc}
\title{A Mechanistic Account of Attention Sinks in GPT-2:\\ One Circuit, Broader Implications for Mitigation}

\author{Yuval Ran-Milo\thanks{Equal contribution.} \and Hila Ofek\footnotemark[1] \and Shahar Mendel \\
  Tel Aviv University \\
  \texttt{\{yuvalmilo, hilaofek1, shaharmendel\}@mail.tau.ac.il}}

\date{}

\begin{document}
\maketitle
\thispagestyle{plain}
\pagestyle{plain}

\begin{abstract}
Transformers commonly exhibit an attention sink: disproportionately high attention to the first position. We study this behavior in GPT-2–style models with learned query biases and absolute positional embeddings. Combining structural analysis with causal interventions, validated across natural-language, mathematical, and code inputs, we find that the sink arises from the interaction among (i) a learned query bias, (ii) the first-layer MLP transformation of the positional encoding, and (iii) structure in the key projection. 
Crucially, each component we identify is individually dispensable: architectures omitting each of them robustly exhibit sinks. This indicates that attention sinks may arise through distinct circuits across architectures. These findings inform mitigation of sinks, and motivate broader investigation into why sinks emerge.
\end{abstract}

\section{Introduction}\label{sec:intro}

Transformers \cite{vaswani2023attentionneed} routinely display an \emph{attention sink}: a persistent tendency to allocate disproportionate attention mass to early (often first) positions independent of semantic content \citep{xiao2023efficient,gu2025when}. The effect has been observed across training stages and hyperparameters \citep{gu2025when,Guo2024ActiveDormantAH}, across model families and datasets \citep{xiao2023efficient}, and under diverse positional encodings—including ALiBi \citep{press2021train}, RoPE \citep{su2021roformer}, and even no explicit positional encodings \citep{gu2025when}. Similar sink-like patterns have also been reported in large multimodal models and vision transformers \citep{Kang2025See,wang2025mirage,Feng2025EDIT}.\footnote{However, some architectures have been reported to have little to no sink \citep{qiu2025gatedattentionlargelanguage, endy-etal-2025-mamba}.}

The practical stakes are significant. Attention sinks can reduce effective context use and lower accuracy \citep{Yu2024Unveiling,Guo2024ActiveDormantAH}, aggravate numerical error and hinder quantization \citep{sun2024massive,lin2024duquant}, obscure interpretability by dominating attention maps \citep{Guo2024ActiveDormantAH}, and complicate streaming and KV-cache strategies \citep{xiao2023efficient}. Analogous effects in vision and multimodal settings waste representational capacity on irrelevant tokens and can be exploited to induce hallucinations \citep{Kang2025See,wang2025mirage,Feng2025EDIT}. Understanding when sinks arise and how to control them is therefore directly relevant for model performance and interpretability.

We study the sink mechanistically in GPT-2–style Transformers with learned query biases and absolute positional embeddings \citep{Radford2019Language}. Combining structural analysis with targeted causal interventions validated across natural-language, mathematical, and code inputs, we tie the first-token sink to three interacting components: learned query bias, first-layer MLP transformation of the positional encoding, and structure in the key projection.\footnote{Our findings connect to and extend a body of work on the origins of attention sinks; see Appendix \ref{app:related_work} for a discussion.} We establish causality by showing that disrupting any component weakens, removes, or relocates the sink. When alternative explanations exist, we ablate them, isolating the causal circuit.

A direct consequence of our analysis is that sinks must arise through distinct mechanisms across architectures: each component we identify is individually dispensable in models that still exhibit sinks. This implies that while attention sinks are robust as a phenomenon, they are not governed by a single universal mechanism. 
Consequently, many post-training mitigations may need to be architecture-specific, and many pre-training architectural interventions may not generalize (see \cref{sec:specificity}).
This highlights the need to understand why the sink emerges as a prerequisite for developing robust pre-training strategies.

  \begin{figure}[t]
    \centering
    \includegraphics[width=1\columnwidth]{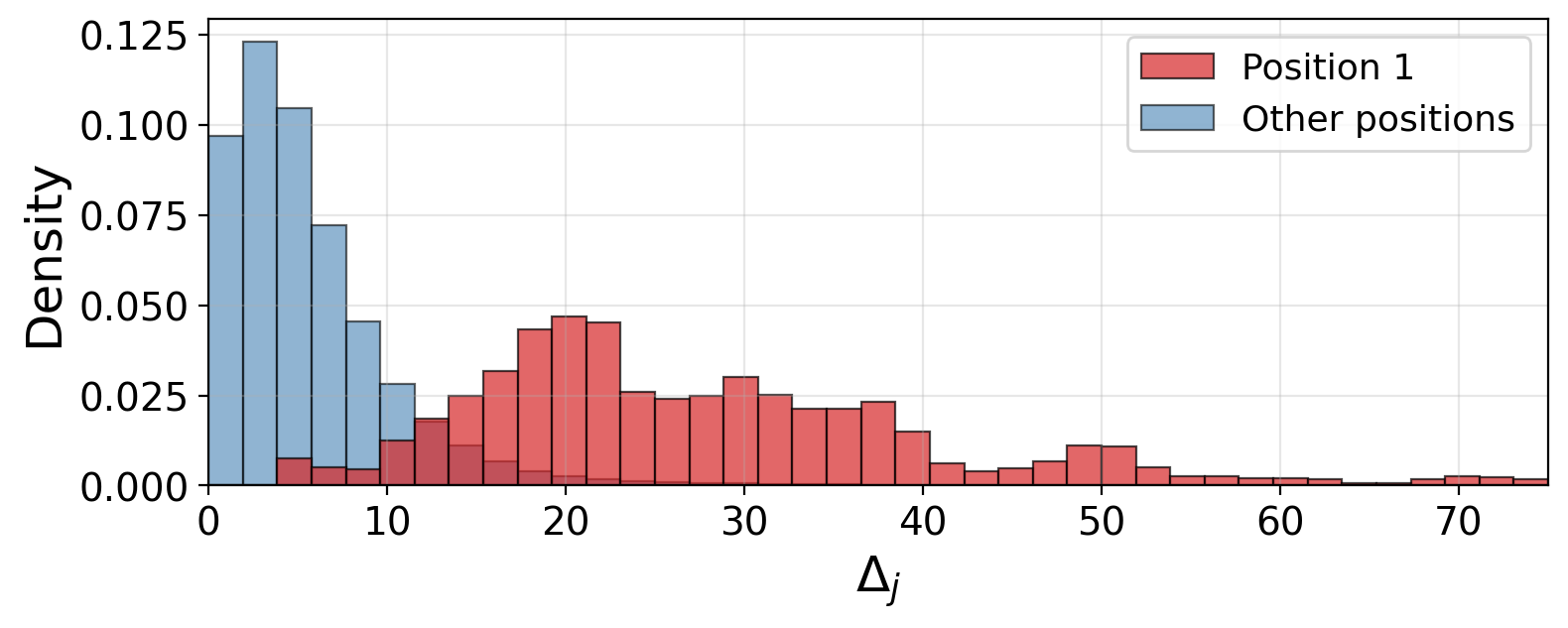}
    \caption{\textbf{The source-agnostic shift $\Delta_{j,h}^{(l)}$ at position~1 is systematically larger than all other positions.} Two overlapping density histograms of the source-agnostic shift $\Delta_{j,h}^{(l)} = b_{Q,h}^{(l)} W_{k,h}^{(l)\top} x_j^{(l)\top}$ (normalized so $\min_j \Delta_{j,h}^{(l)} = 0$; see \cref{sec:delta_analysis}), pooled across all heads $h$ and layers $l \in [4, 11]$ (see \cref{sec:setup}). 
    The distribution of scores for the first position ($j=1$, red) is clearly separated from and systematically larger than the distribution for all other positions ($j>1$, blue). The $x$-axis is truncated for clarity; see full histogram in Appendix~\ref{app:full_delta}.}
    \label{fig:delta_percentiles}
  \end{figure}

\vspace{-0.3em}
\section{Preliminaries}

\vspace{-0.4em}
\subsection{Attention mechanism}\label{sec:attn_mechanism}
We use $D$ for model dimension, $H$ for number of heads, $D_h$ for per-head dimension, and $L$ for number of layers. We index layers by $l$, heads by $h$, and sequence positions by $i$ or $j$. The representation at layer $l$, position $i$ is $x_i^{(l)} \in \mathbb{R}^D$. Within each head $h$, queries, keys, and values are $q_{i,h}^{(l)} = x_i^{(l)} W_{q,h}^{(l)} + b_{Q,h}^{(l)}$, $k_{j,h}^{(l)} = x_j^{(l)} W_{k,h}^{(l)} + b_{K,h}^{(l)}$, and $v_{j,h}^{(l)} = x_j^{(l)} W_{v,h}^{(l)} + b_{V,h}^{(l)}$, where $W_{q,h}^{(l)}, W_{k,h}^{(l)}, W_{v,h}^{(l)} \in \mathbb{R}^{D \times D_h}$ and biases $b_{Q,h}^{(l)}, b_{K,h}^{(l)}, b_{V,h}^{(l)} \in \mathbb{R}^{D_h}$. The pre-softmax attention score from position $i$ to position $j$ in head $h$ is $s_{i \to j,h}^{(l)} = q_{i,h}^{(l)} (k_{j,h}^{(l)})^\top$. The attention weights are $\alpha_{i \to j,h}^{(l)} = \mathrm{softmax}_j\!\bigl(s_{i \to j,h}^{(l)} / \sqrt{D_h}\bigr)$ over valid positions $j \le i$, and the output for position $i$ is $o_{i,h}^{(l)} = \sum_j \alpha_{i \to j,h}^{(l)}\, v_{j,h}^{(l)}$; head outputs are concatenated and projected by $W_o^{(l)} \in \mathbb{R}^{D \times D}$.
\footnote{We adopt the usual transformer convention $D = H D_h$.}

\subsection{Positional encoding}
Attention layers are not inherently order-aware; apart from masking-induced structure, they are invariant to input permutations. To address this, Transformers incorporate positional information through various schemes (\citet{su2021roformer}, \citet{press2021train}). 
GPT-2 uses learned absolute positional encodings: a set of trainable vectors $\{p_i\}_{i=1}^{N} \subset \mathbb{R}^{D}$, where $i$ is the token position and $N$ is the maximal sequence length. These are added to token embeddings $e_i$: $x_i^{(0)} = e_i + p_i$.

\begin{figure}[t]
  \centering
  \includegraphics[width=1\columnwidth]{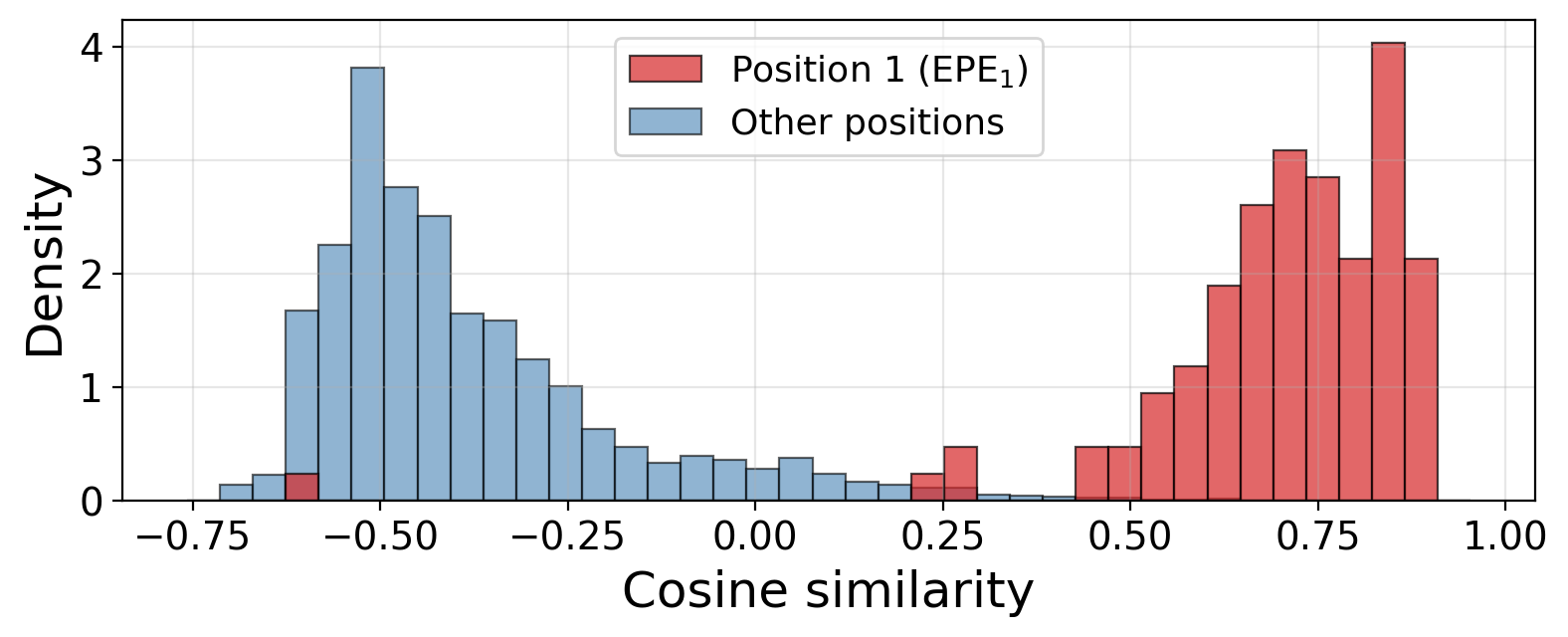}
  \caption{\textbf{The first-position key projection $\mathrm{EPE}_1 W_{k,h}^{(l)}$ is uniquely aligned with the query bias.} Two overlapping density histograms of the per-head cosine similarity between $b_{Q,h}^{(l)}$ and $\mathrm{EPE}_j W_{k,h}^{(l)}$, pooled across all heads $h$ and layers $l \in [4, 11]$. The first-position distribution ($j{=}1$, red) is concentrated at high positive values, indicating strong alignment with the query bias, while the distribution for all other positions ($j{>}1$, blue) is negatively aligned.}
  \label{fig:obs2_layer10}
\end{figure}

\subsubsection{Effective positional encoding (EPE)}
We define the \emph{effective positional encoding} (EPE) for position $i$ as $\mathrm{EPE}_i = \mathrm{MLP}^{(1)}(p_i) + p_i \in \mathbb{R}^D$, where $\mathrm{MLP}^{(1)}$ is the first layer's feed-forward network. We call this ``effective'' because it roughly captures the net positional signal that the first layer's MLP writes into the residual stream
\footnote{This equals \emph{exactly} the positional contribution if we simplify the first layer to only a linear MLP (removing its nonlinearity), omitting attention and removing normalization.}:
 as we verify in \cref{sec:epe_validation}, adding $\mathrm{EPE}_i$ to the first layer's output produces roughly the same result as adding $p_i$ before the first layer and then applying the first layer's MLP.

\section{Methodology and Results}
We first describe the mechanism underlying attention sinks in models with learned query biases and absolute positional encodings in \cref{sec:mechanism}. We then provide empirical evidence through four analyses in \cref{sec:delta_analysis,sec:epe_validation,sec:epe_alignment,sec:wk_structure} and confirm causality through targeted interventions in \cref{sec:interventions}.

\subsection{The Mechanism behind Attention Sinks}\label{sec:mechanism}
Expanding the attention score $s_{i \to j,h}^{(l)}$ (layer $l$, head $h$, source position $i$, target position $j$) from \cref{sec:attn_mechanism} by substituting the expressions for $q_{i,h}^{(l)}$ and $k_{j,h}^{(l)}$ gives $s_{i \to j,h}^{(l)} = \bigl(x_i^{(l)} W_{q,h}^{(l)}\bigr)\bigl(x_j^{(l)} W_{k,h}^{(l)}\bigr)^\top + \bigl(x_i^{(l)} W_{q,h}^{(l)}\bigr) b_{K,h}^{(l)\top} + b_{Q,h}^{(l)} \bigl(x_j^{(l)} W_{k,h}^{(l)}\bigr)^\top + b_{Q,h}^{(l)} b_{K,h}^{(l)\top}$.
The third term, $\Delta_{j,h}^{(l)} \triangleq b_{Q,h}^{(l)} W_{k,h}^{(l)\top} x_j^{(l)\top}$, is a token-specific, \emph{source-agnostic} 
shift: it raises or lowers the score toward target position $j$ identically for every source position $i$. We find that this term for the first 
token, $\Delta_{1,h}^{(l)}$, is conspicuously large across heads in most deep layers, creating a strong prior to attend to position~1.

To understand why $\Delta_{1,h}^{(l)}$ is so large, recall that the effect of adding $p_1$ to the first token and passing it through the first layer is roughly equivalent to adding $\mathrm{EPE}_1$ directly to the residual stream at position~1 (as verified in \cref{sec:epe_validation}). $\mathrm{EPE}_1$ has very large absolute values on a small set of coordinates---the phenomenon of \textit{massive 
activations} \citep{sun2024massive}---which are exactly those coordinates where $b_{Q,h}^{(l)} W_{k,h}^{(l)\top}$ has the largest magnitude in 
almost all layers. This co-adaptation allows $\mathrm{EPE}_1$ to dramatically amplify $\Delta_{1,h}^{(l)}$ across heads, yielding an attention sink at position~1.

\subsection{Empirical Validation}

  \begin{figure}[t]
    \centering
    \includegraphics[width=1\columnwidth]{figures/epe_exp.png}
    \caption{\textbf{EPE closely tracks the net positional signal at every position.} Per-position cosine similarity between $\mathrm{EPE}_i$ and the net added positional signal $N_i$ (see \cref{sec:epe_validation}). The shaded band spans the 10th--90th percentile across our dataset; the line marks the median. The similarity exceeds $0.82$ at every position and reaches $\boldsymbol{>}0.99$ at position~1, confirming that $\mathrm{EPE}_1$ (the key driver of the sink) is an accurate proxy for the actual positional contribution.}
    \label{fig:epe_exp}
  \end{figure}

We validate our proposed mechanism through four complementary analyses, followed by causal interventions that confirm the necessity of each component described in \cref{sec:mechanism}. The model and evaluation dataset are described in \cref{sec:setup}. In \cref{sec:delta_analysis} we show that $\Delta_{1,h}^{(l)}$ is conspicuously large relative to other positions across layers and heads. In \cref{sec:epe_validation} we verify that the EPE accurately captures the net positional signal written by the first layer's MLP. We then investigate the underlying cause of the large shift and show in \cref{sec:epe_alignment} that $\mathrm{EPE}_1 W_{k,h}^{(l)}$ is strongly aligned with $b_{Q,h}^{(l)}$ across heads and layers. In \cref{sec:wk_structure} we establish that $\mathrm{EPE}_1$ exhibits massive activations precisely at coordinates where the bias projection $b_{Q,h}^{(l)} W_{k,h}^{(l)\top}$ has high magnitude. Finally, in \cref{sec:interventions} we use causal interventions to verify that disrupting any component weakens or removes the sink while transplanting it transfers the sink to a new position.

\subsubsection{Setup}\label{sec:setup}
All experiments are conducted on the 124M-parameter GPT-2 \citep{Radford2019Language}\footnote{Pretrained checkpoint from the Hugging Face Hub: \url{https://huggingface.co/openai-community/gpt2}; License: MIT.}.
We evaluate on a dataset of 300 examples: 100 randomly sampled from each of SST-2 \citep{socher-etal-2013-recursive} (natural language), GSM8K \citep{cobbe2021training} (mathematical reasoning), and HumanEval \citep{chen2021evaluating} (code generation). Only examples with at least 40 tokens are kept; all selected examples are then truncated to exactly 40 tokens so that every input has the same sequence length. This standardization ensures that no dataset disproportionately influences aggregate statistics due to varying input lengths; we find that results are qualitatively equivalent for other length choices.
All per-head analyses are reported over layers~4--11 (of~12), excluding the first three layers, and the last layer, where attention sinks are known to be weaker \cite{yu2024unveilingharnessinghiddenattention}.
Code and data are available at \url{https://github.com/YuvMilo/MechanisticAccountofSinks}.

\subsubsection{Source-Agnostic Shift Analysis}
\label{sec:delta_analysis}
First, we verify that $\Delta_{1,h}^{(l)}$ is anomalously large relative to other positions. We compute $\Delta_{j,h}^{(l)}$ for every head $h$ and position $j$ across our evaluation dataset. 
Because softmax attention is invariant to adding a constant across all target positions, we re-center $\Delta$ by subtracting $\min_j \Delta_{j,h}^{(l)}$ from all positions for each (sentence, layer, head) slice, setting the smallest value to zero; this does not affect the model's output and makes magnitudes comparable across slices.
\Cref{fig:delta_percentiles} shows two overlapping density histograms of $\Delta_{j,h}^{(l)}$ across all heads $h$ and layers $l \in [4, 11]$: one for the first position ($j{=}1$, red) and one for all other positions ($j{>}1$, blue). The two distributions are clearly separated, with the first-position distribution shifted far to the right, confirming that $\Delta_{1,h}^{(l)}$ is indeed anomalously large and creates a strong, consistent prior to attend to position~1.
Later, in \cref{sec:interventions}, we show that nullifying $b_Q$ alone---effectively zeroing out this term in every attention score---is sufficient to drastically diminish the sink.

\subsubsection{EPE Captures the Net Positional Contribution}
\label{sec:epe_validation}
Our mechanism relies on $\mathrm{EPE}_i$ faithfully representing the net positional signal written into the residual stream by the first layer's MLP. To verify this, we define the \emph{net added positional signal} $N_i \coloneq R_i - O_i$, where $R_i = x_i^{(0)} + \mathrm{MLP}^{(1)}(x_i^{(0)})$ is the MLP output given an input with positional encoding, and $O_i = e_i + \mathrm{MLP}^{(1)}(e_i)$ is the output for the same token embedding without positional encoding ($e_i = x_i^{(0)} - p_i$). $N_i$ is thus the incremental change to the residual stream attributable to the positional encoding. We then measure the cosine similarity between $\mathrm{EPE}_i$ and $N_i$ across our dataset (see \cref{sec:setup}) and all positions. \Cref{fig:epe_exp} shows that the median similarity exceeds $0.82$ at every position. Crucially, for position~1 (the sink position), the cosine similarity exceeds $0.99$. 

Furthermore, we conducted an additional experiment comparing the cosine similarity of $\mathrm{EPE}_1$ and the added positional signal of the entire first layer (including normalization and first layer attention) in a similar fashion, and found the cosine similarity median to be $0.76$ (90th and 10th percentiles are 0.8 and 0.67 respectively).
\footnote{This comparison is meaningful only at position~1, which under causal masking attends solely to itself and thus carries a pure positional signal; at other positions, attention mixes representations from preceding tokens, confounding the positional contribution.} This implies that adding $p_1$ to the first token and passing it through the first layer is roughly equivalent to adding $\mathrm{EPE}_1$ directly to the residual stream at position~1, implying that $\mathrm{EPE}_1$ accurately captures the positional signal. 

\begin{figure}[t]
  \centering
  \includegraphics[width=\columnwidth]{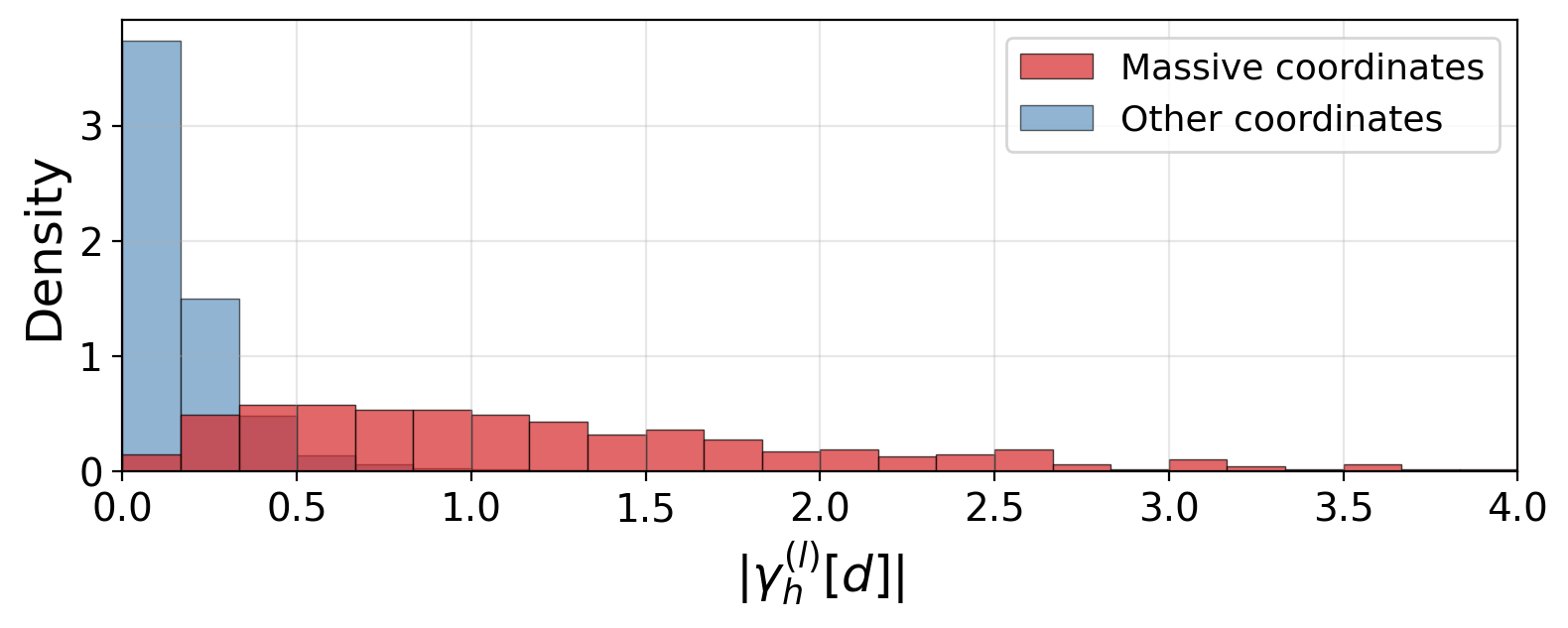}
  \caption{\textbf{$\mathrm{EPE}_1$ is large exactly where the bias projection is large.} Two overlapping density histograms of $|\gamma_h^{(l)}[d]|$, across all heads $h$ and layers $l \in [4, 11]$: massive-activation coordinates of $\mathrm{EPE}_1$ (red) versus all other coordinates (blue). The $x$-axis is truncated for clarity; see full histogram in Appendix~\ref{app:coor_align}.}
  \label{fig:coord_hist_trunc}
\end{figure}

\subsubsection{EPE-Bias Projection Alignment}
\label{sec:epe_alignment}
Having established the magnitude of $\Delta_{1,h}^{(l)}$, we investigate its underlying cause. Since $x_1^{(l)}$ contains both token and positional information, it remains to disentangle which of the two is responsible for the large $\Delta_{1,h}^{(l)}$. To that end, we compute the cosine similarity between $b_{Q,h}^{(l)}$ and $\mathrm{EPE}_j W_{k,h}^{(l)}$ for every head $h$, position $j$, and layer $l$. \Cref{fig:obs2_layer10} shows two overlapping density histograms pooled over layers~4--11 and all heads: the first-position distribution ($j{=}1$, red) is tightly concentrated at high positive values, indicating that the first-position key projection $\mathrm{EPE}_1 W_{k,h}^{(l)}$ is strongly aligned with the query bias, while the distribution for all other positions ($j{>}1$, blue) is negatively aligned.
Since $\mathrm{EPE}_1$ is the only position positively aligned with the query bias, its addition to the residual stream (see \cref{sec:epe_validation}) directly enlarges $\Delta_{1,h}^{(l)}$, attributing the outsized shift to the positional signal.

\subsubsection{Coordinate-Level Structural Analysis}
\label{sec:wk_structure}
Since $\Delta_{j,h}^{(l)} = b_{Q,h}^{(l)} W_{k,h}^{(l)\top} x_j^{(l)\top}$, the large $\Delta_{1,h}^{(l)}$ requires $b_{Q,h}^{(l)} W_{k,h}^{(l)\top}$ to have high magnitude at the same coordinates where $\mathrm{EPE}_1$ is large. We verify this by comparing, for each head $h$, the bias projection magnitude $\gamma_h^{(l)}[d] \coloneq |b_{Q,h}^{(l)} \cdot W_{k,h}^{(l)}[:,d]|$ at the massive-activation coordinates of $\mathrm{EPE}_1$ (identified in Appendix \ref{app:massive_activations_in_ppe}) against all other coordinates. \Cref{fig:coord_hist_trunc} shows two overlapping density histograms of $\gamma_h^{(l)}[d]$ across all heads $h$ and layers $l \in [4, 11]$: massive-activation coordinates (red) are systematically larger than all other coordinates (blue), confirming the coordinate-level co-adaptation that amplifies $\Delta_{1,h}^{(l)}$

\begin{figure*}[t!]
  \includegraphics[width=\textwidth]{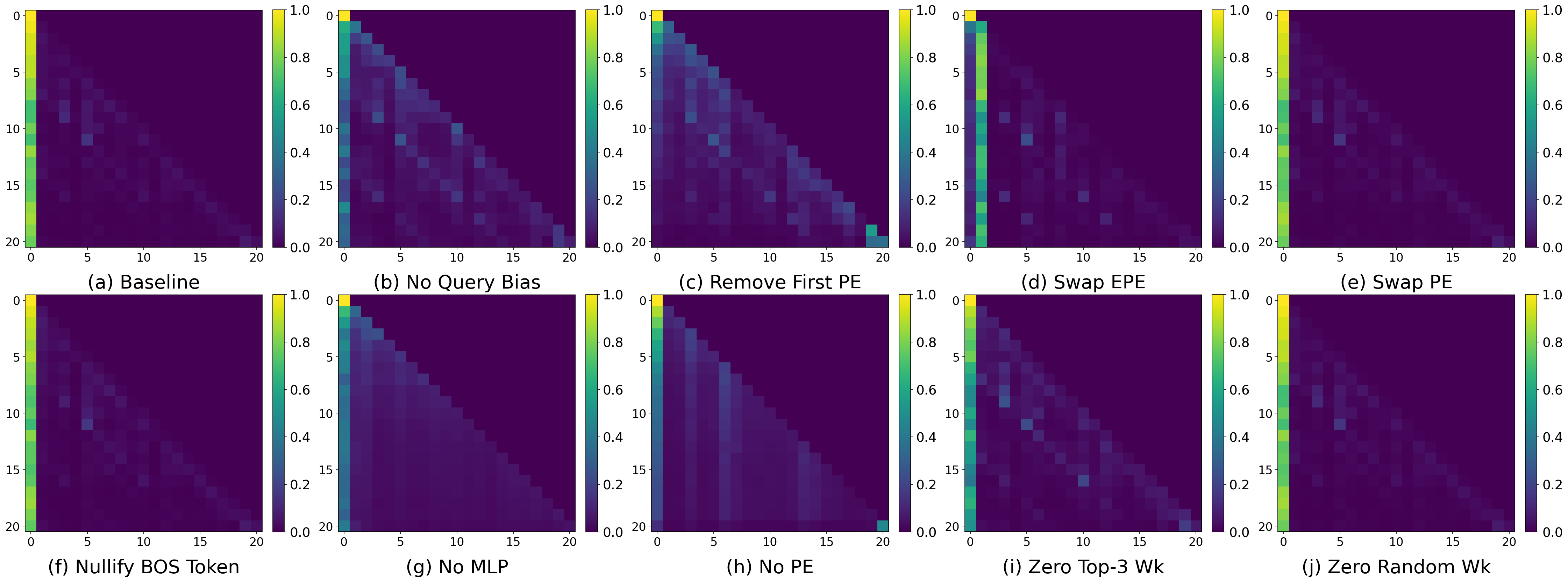}
  \caption{\textbf{Disrupting any component of the $b_Q$--$\mathrm{EPE}_1$--$W_k$ pathway significantly diminishes the sink.} Head-averaged attention maps (layers~4--11) under all ten interventions for a single representative sentence.  Each panel shows attention weights from query positions (rows) to key positions (columns).  The first-position sink (bright first column) persists when the pathway is left intact~(a,\,e,\,f,\,j), but is significantly diminished when any component is disrupted~(b,\,c,\,d,\,g,\,h,\,i). See \cref{tab:interventions} for aggregate statistics across our dataset.}
  \label{fig:interventions_comparison}
\end{figure*}

\subsubsection{Causal Interventions}
\label{sec:interventions}
To establish causality beyond correlation, we perform ten targeted interventions during forward passes, testing the necessity of components by removing them and sufficiency by transplanting them. Where alternative explanations exist, we include control interventions to ablate them.
  We quantify each intervention's effect using the \textbf{BOS-attention metric}: the average attention weight assigned to position~1 by tokens in the second half of the sequence,\footnote{We restrict to the second half so that early positions, whose attention is concentrated on position~1 simply because few alternatives are available, do not inflate the metric.} averaged over all heads and layers~4--11 (see \cref{sec:setup}).  Results are in \cref{tab:interventions}. \Cref{fig:interventions_comparison} illustrates the effect of each intervention on the attention map for a single representative sentence.\footnote{``It was the best of times, it was the worst of times, it was the age of wisdom.'' (from \emph{A Tale of Two Cities} by Charles Dickens, 1859).}

\begin{table}[t]
  \centering
  \small
  \begin{tabular}{llcc}
    \hline
    & \textbf{Intervention} & \textbf{BOS Attn} & \textbf{\% Base} \\
    \hline
    (a) & Baseline                    & $.563{\scriptstyle\pm.004}$ & $100.0$  \\
    (b) & Nullify $b_Q$               & $.251{\scriptstyle\pm.003}$ & $44.7$   \\
    (c) & Remove First PE             & $.017{\scriptstyle\pm.001}$ & $3.0$    \\
    (d) & Swap EPE                    & $.035{\scriptstyle\pm.002}$ & $6.2$    \\
    (e) & Swap PE                     & $.557{\scriptstyle\pm.006}$ & $99.0$   \\
    (f) & Nullify BOS Token           & $.561{\scriptstyle\pm.004}$ & $99.7$   \\
    (g) & No MLP                      & $.283{\scriptstyle\pm.009}$ & $50.3$   \\
    (h) & No PE                       & $.099{\scriptstyle\pm.015}$ & $17.5$   \\
    (i) & Zero Top-3 $W_k$            & $.367{\scriptstyle\pm.004}$ & $65.2$   \\
    (j) & Zero Random $W_k$           & $.563{\scriptstyle\pm.004}$ & $100.0$  \\
    \hline
  \end{tabular}
  \caption{\textbf{Each component of the sink circuit is necessary.} BOS-attention metric (mean $f\pm$ 2\,SE) across our dataset, averaged over all heads and layers~4--11.  ``\%~Base'' is the fraction of the baseline~(a) score retained.  Disrupting any elements of the $b_Q$--$\mathrm{EPE}_1$--$W_k$ pathway (b--d,\,g--i) substantially reduces the sink; controls (e,\,f,\,j) do not.}
  \label{tab:interventions}
\end{table}

\vspace{-5pt}
\begin{itemize}[leftmargin=*,itemsep=1pt]
  \item \textbf{Baseline.} No intervention; the attention sink is clearly present.
  \item \textbf{Nullify $b_Q$.} We set $b_Q{=}0$ in every layer; the sink substantially diminishes, showing that $b_Q$ is necessary for the large first-token contribution.
  \item \textbf{Remove First PE.} We replace $\mathrm{PE}_1$ with $\mathrm{PE}_2$ at position~1; the sink nearly vanishes. Together with the BOS-nullification control~(f) below, this shows that the positional encoding, not the BOS token identity, drives the sink.
  \item \textbf{Swap EPE.} We perform a swap of the EPE component between positions~1 and~2\footnote{Results are qualitatively the same when swapping position~1 and any other position, not necessarily 2.} in the first layer's MLP output (technical details in \cref{app:swap_details}). The sink nearly vanishes, demonstrating that the MLP-transformed positional signal controls sink location.
  \item \textbf{Swap PE.} Same swap as above but using the raw positional embeddings instead of the EPE (details in \cref{app:swap_details}). The sink persists, showing that downstream layers rely specifically on the MLP-processed signal, not the raw positional embedding.
  \item \textbf{Nullify BOS Token.} Zeroing the BOS token embedding before adding positional signals leaves the sink intact, ruling out BOS token identity as a driver.
  \item \textbf{No MLP.} Skipping all MLP blocks substantially reduces the sink,\footnote{We disable all MLPs rather than only the first because we observe that later MLPs also process the positional encoding in a qualitatively similar manner, though more weakly.} consistent with the first-layer MLP's role in amplifying the positional signal into the EPE.
  \item \textbf{No PE.} Removing all positional embeddings greatly diminishes the sink, confirming that positional information is the primary driver.
  \item \textbf{Zero Top-3 $W_k$.} Zeroing the three $W_k$ columns at the massive-activation coordinates of $\mathrm{EPE}_1$ reduces the sink, confirming that those coordinates are the mechanism through which $\mathrm{EPE}_1$ amplifies $\Delta_{1,h}^{(l)}$.
  \item \textbf{Zero Random $W_k$.} Zeroing three random $W_k$ columns has nearly no effect, confirming that the result above is specific to the massive-activation coordinates of $\mathrm{EPE}_1$.
\end{itemize}

\section{Conclusions}\label{sec:specificity}

Through fine-grained structural analysis and targeted causal interventions, validated across natural-language, mathematical, and code inputs, we identified the circuit behind attention sinks in GPT-2: the interaction of a learned query bias ($b_Q$), the first-layer MLP's transformation of the positional encoding, and coordinate-level structure in the key projection ($W_k$). Disrupting any single component significantly weakens the sink, while alternative explanations are ablated.

Crucially, each component of this circuit is individually dispensable:
many architectures omit $b_Q$ entirely yet still display strong attention sinks \citep{gu2025when,xiao2023efficient}, 
sinks persist even in Transformers without any MLP blocks \citep{hong2025variance}, and sinks persist in Transformers not using any positional encodings \citep{gu2025when}.
 The sink therefore arises through distinct circuits across architectures\footnote{Other architectures may construct similar mechanisms using different tools (e.g., a constant-feature slice of $W_Q$ could substitute for an absent $b_Q$). Our claim is that the \emph{specific} parameter-level pathway we identify does not carry over.}---suggesting it reflects a structural computational necessity that optimization reliably learns using whatever architectural components are available. \footnote{This view is supported by works showing that sinks are provably necessary for certain attention computations \citep{ranmilo2026attentionsinksprovablynecessary}, and that they serve a functional role in mitigating over-mixing \citep{Barbero2025Why}.}

This implies that while attention sinks are robust as a phenomenon, they are not governed by a single universal circuit. Consequently, many pre-training architectural modifications aimed at preventing sinks may prove insufficient, as optimization can potentially reconstruct an equivalent sink using alternative components. Likewise, many post-training mitigations may need to be architecture-specific: targeting a particular component will succeed only if it forms part of the active sink-implementing pathway in that specific model. 
We hope our causal analysis serves as a first step toward designing architecture-aware mitigations, and motivates further investigation into why the sink emerges as a prerequisite for robust pre-training strategies.

\section{Limitations}

\subsection{Scope across architectures and scales}
Our analyses focus on a GPT-2–style model with learned query biases and absolute positional encodings. The broader Transformer ecosystem includes architectures that omit such biases or use alternative positional schemes (e.g., RoPE, ALiBi). While we do show that the specific components constituting the GPT-2 circuit are absent or altered in those settings, we leave the investigation of what mechanisms do form in those models to future work. In addition, GPT-2 is small by contemporary standards; with scale, the mechanism could strengthen, fragment into multiple pathways, or be replaced by different circuits.

\subsection{Learning dynamics}
We provide a post-hoc, static analysis of a trained checkpoint. We do not track when the circuit emerges during pre-training, which gradients give rise to it, or whether intermediate snapshots exhibit qualitatively different pathways. We believe our static analysis could inform future work researching the emergence of the attention sink mechanism.

\subsection{Secondary contributors}
Our interventions reduce the sink substantially but do not eliminate it entirely: nullifying $b_Q$ or zeroing the massive-activation rows of $W_k$ leaves a residual sink. This indicates that secondary contributors beyond the three main components we isolate also play a role. Identifying them is a natural extension of our work which we leave to future work.

\subsection{Mechanism vs. function}
Our contribution is mechanistic: we explain \emph{how} an attention sink can be implemented in the studied architecture. We do not claim a definitive functional rationale for \emph{why} such a sink is beneficial or harmful across tasks. Establishing the downstream utility or cost of the sink, and the conditions under which it is selected by optimization, is left for future work.

\section*{Acknowledgments}
I thank Amit Elhelo and Daniela Gottesman for illuminating discussions. Special thanks to my advisor Nadav Cohen for his guidance and mentorship.  We used AI assistance for writing and code development. This work was supported by the European Research Council (ERC) grant NN4C 101164614, a Google Research Scholar Award, a Google Research Gift, Meta, the Yandex Initiative in Machine Learning, the Israel Science Foundation (ISF) grant 1780/21, the Tel Aviv University Center for AI and Data Science, the Adelis Research Fund for Artificial Intelligence, Len Blavatnik and the Blavatnik Family Foundation, and Amnon and Anat Shashua.

\bibliography{custom}

\appendix

\section{Further Experiments}

\subsection{Full Source-Agnostic Shift Histogram}\label{app:full_delta}

\begin{figure}[t]
  \centering
  \includegraphics[width=\columnwidth]{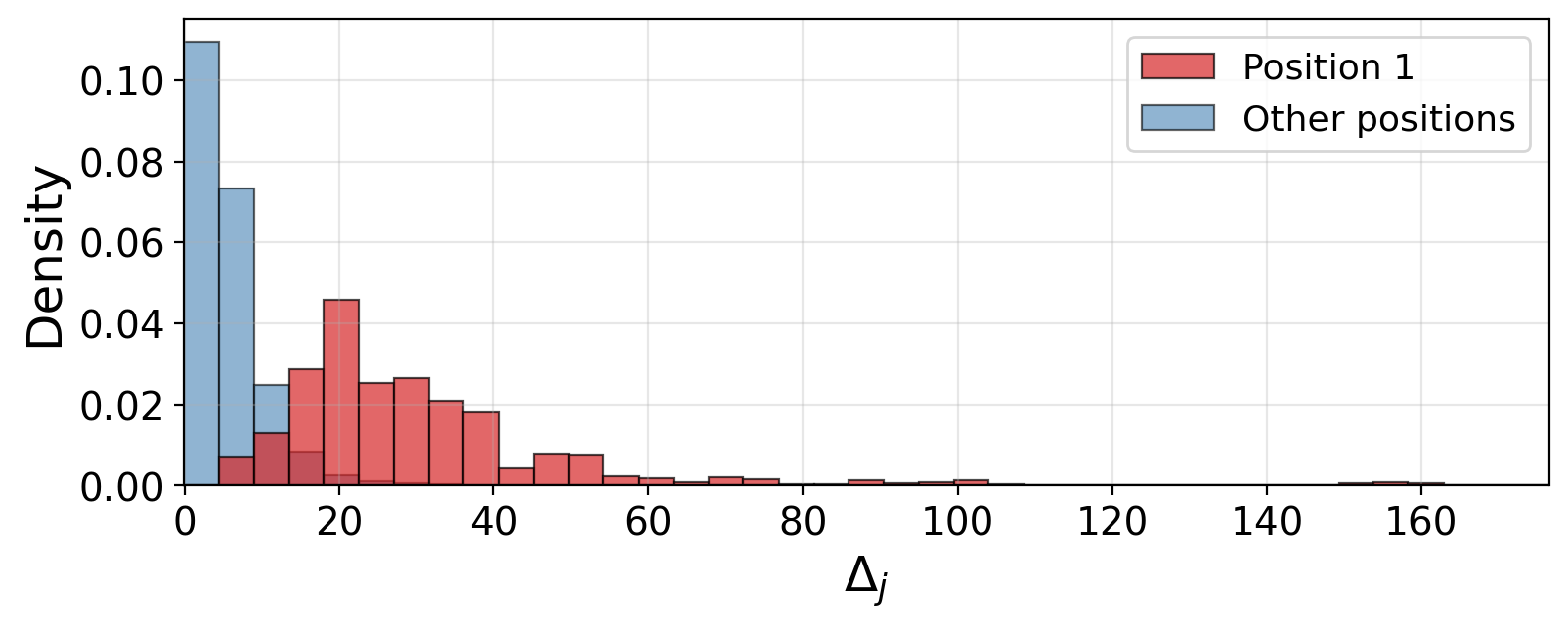}
  \caption{\textbf{Full (non-truncated) histogram of the source-agnostic shift $\Delta_{j,h}^{(l)}$.} Same data as \cref{fig:delta_percentiles} with no axis restriction. Scores are normalized per (sentence, layer, head) slice so the minimum is zero. The first-position distribution (red) extends to a long right tail that lies well beyond the range of all other positions (blue).}
  \label{fig:full_delta_histogram}
\end{figure}

\Cref{fig:full_delta_histogram} shows the full (non-truncated) version of \cref{fig:delta_percentiles}.

\subsection{Identifying Massive Activations in First-Position EPE}\label{app:massive_activations_in_ppe}

\begin{figure}
  \includegraphics[width=\columnwidth]{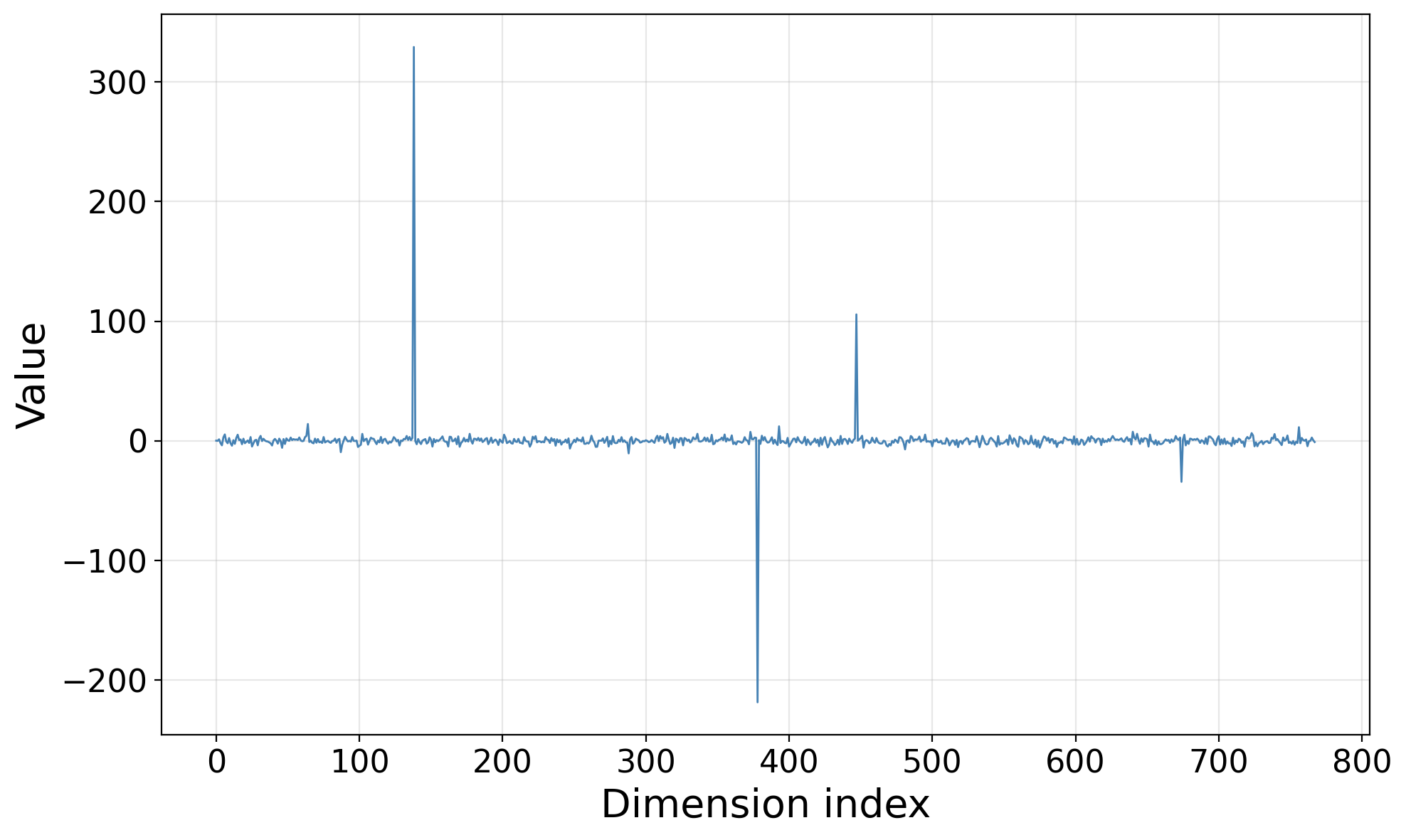}
  \caption{\textbf{A few coordinates of $\mathrm{EPE}_1$ have massive activations.} Coordinate values of $\mathrm{EPE}_1$. Most coordinates are near zero; dims 138, 378, and 447 exhibit extremely large magnitudes.}
  \label{fig:appendix_massive_activations}
\end{figure}

This section explains how we identify coordinates with unusually large absolute values in $\mathrm{EPE}_1$. We select coordinates whose absolute values exceed the mean absolute value by at least three standard deviations; in our model this criterion selects indices 138, 378 and 447. Elements at these indices are clear outliers, each more than 15 standard deviations away from the mean. Each such selected dimension exhibits the coordinate-level phenomenon described in \cref{sec:wk_structure} (i.e., large $|\gamma_h^{(l)}[d]|$ and a strong contribution to the source-agnostic shift). See \cref{fig:appendix_massive_activations} for a visualization of $\mathrm{EPE}_1$. It is clear visually that coordinates 138, 378, and 447 have conspicuously larger norms than other indices. 

\subsection{Full Coordinate-Level Alignment Histogram} \label{app:coor_align}

\begin{figure}[t]
  \centering
  \includegraphics[width=\columnwidth]{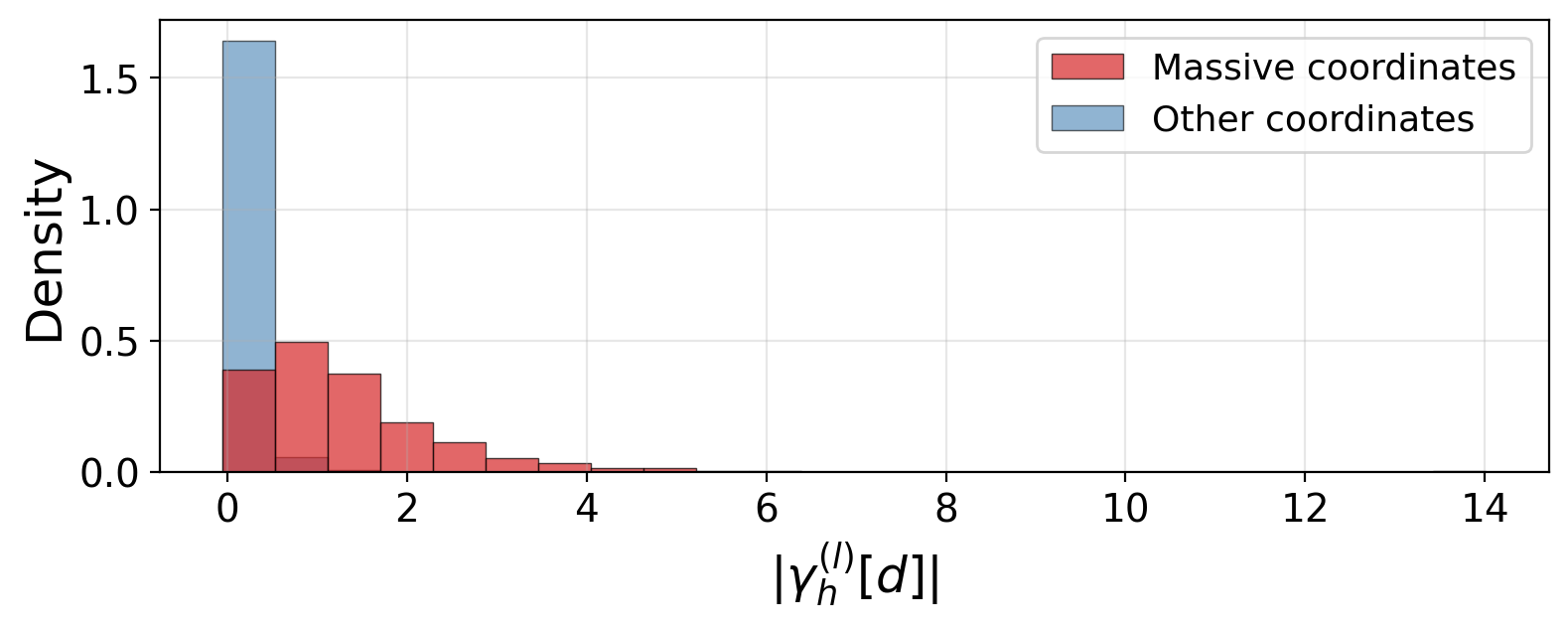}
  \caption{\textbf{Full (non-truncated) histogram of $|\gamma_h^{(l)}[d]|$.} Same data as \cref{fig:coord_hist_trunc} with no axis restriction. The massive-activation coordinates of $\mathrm{EPE}_1$ (red) extend to a long right tail that lies well beyond the range of all other coordinates (blue).}
  \label{fig:full_coord_histogram}
\end{figure}

\Cref{fig:full_coord_histogram} shows the full (non-truncated) version of \cref{fig:coord_hist_trunc}.

\section{Related Work}\label{app:related_work}

Our work continues a growing line of efforts to understand attention sinks and related extreme-token phenomena, including empirical, mechanistic, and functional analyses by \citet{xiao2023efficient}, \citet{Yu2024Unveiling}, \citet{gu2025when}, \citet{Guo2024ActiveDormantAH}, \citet{sun2024massive}, \citet{lindnielsen2024spectral}, \citet{Barbero2025Why}, \citet{ruscio2025geometric}, \citet{qiu2026unified}, \citet{sun2026spikesparse}, \citet{hong2025variance}, \citet{dellano2026compression}, \citet{ranmilo2026attentionsinksprovablynecessary}, and \citet{dial2025bos}.
Of particular relevance are the following works, which resonate most closely with our findings; below we clarify how our analysis relates to and extends each of them.

\citet{gu2025when} show that attention sinks emerge during pretraining, that their location depends on factors such as the loss and the data distribution, and that sinks behave more like key-bias-like extra scores than semantically meaningful content.
Our analysis is complementary but substantially more fine-grained.
Rather than stopping at the conclusion that BOS semantics are not the right explanation, we show what specifically replaces that explanation in GPT-2: the sink is driven by the positional signal at position~1 after it is transformed by the first-layer MLP.
Concretely, we separate BOS token signal from positional information with a direct intervention: nullifying the BOS token leaves the sink essentially unchanged, whereas removing the first positional embedding largely destroys it; moreover, swapping the EPE strongly affects the sink, while swapping the raw positional embedding does not.

\citet{sun2024massive} identify massive activations as outlier coordinates whose values are largely input-invariant, function as indispensable bias terms, and concentrate attention on the tokens that carry them.
We build directly on this observation, but add the missing upstream explanation for the GPT-2 sink: in our setting, the relevant massive activations are not merely correlated with the sink token, but are generated from the first positional encoding through the first-layer MLP.
Put differently, we do not only show that massive activations matter for sinks; we trace the sink-relevant massive coordinates back to a concrete source, namely the MLP-transformed positional signal at position~1.

\citet{dial2025bos} is especially close in spirit to our work.
He shows in GPT-2-small that a prominent sink-associated coordinate grows in an early MLP layer, tracing its progression through the MLP computation.
Our analysis sharpens this in two ways.
First, we formalize the relevant object as the effective positional encoding ($\mathrm{EPE}$), and verify quantitatively that it closely tracks the net positional signal written by the first layer.
Second, we show causally that the crucial object is not merely an early-MLP feature, but specifically the MLP-processed positional signal: swapping the EPE sharply disrupts the sink, whereas swapping the raw positional embedding does not.

\citet{dial2025bos} also shows that downstream queries tend to align with the key-weight direction associated with the sink-related massive coordinate.
We extend this by identifying the precise parameter-level mechanism through which this alignment affects attention scores.
The key object is the alignment between $\mathrm{EPE}_1 W_{k,h}^{(l)}$ and the learned query bias $b_{Q,h}^{(l)}$, which induces a large source-agnostic score shift $\Delta_{1,h}^{(l)}$ toward position~1, connecting the observed alignment to a concrete sink-forming circuit rather than treating it as an isolated geometric regularity.

\subsection{EPE and PE Swap Details}\label{app:swap_details}

Interventions~(d) and~(e) test whether the sink is controlled by the direction of the EPE (or raw PE) at position~1. Let $m_j$ denote the first-layer MLP output at position~$j$, and let $\hat{u}_j = u_j / \lVert u_j \rVert$ be the unit-norm direction to swap ($u_j = \mathrm{EPE}_j$ for intervention~d; $u_j = p_j$ for intervention~e). We compute $\alpha = m_1 \cdot \hat{u}_1$, the projection of $m_1$ onto the position-1 direction, project this component out of $m_1$, and replace it with the position-2 direction. We apply the mirror operation at position~2 using the same coefficient:
\[
m_1 \leftarrow m_1 - \alpha\,\hat{u}_1 + \alpha\,\hat{u}_2, \quad
m_2 \leftarrow m_2 + \alpha\,\hat{u}_1 - \alpha\,\hat{u}_2.
\]
Using the same scalar $\alpha$ for the removed and added components makes the swap magnitude-preserving: only the positional direction changes, not its norm.

\end{document}